\documentclass{article}

\usepackage{microtype}
\usepackage{graphicx}
\usepackage{subcaption}
\usepackage{booktabs} 

\usepackage{hyperref}

\usepackage[preprint]{icml2026}

\usepackage{amsmath}
\usepackage{amssymb}
\usepackage{mathtools}
\usepackage{amsthm}

\usepackage[capitalize,noabbrev]{cleveref}

\usepackage{enumitem}

\theoremstyle{plain}

\theoremstyle{definition}

\theoremstyle{remark}

\usepackage[textsize=tiny]{todonotes}

\icmltitlerunning{No Generation without Representation}

\begin{document}

\twocolumn[
  \icmltitle{No Generation without Representation: Efficient Causal Protein Language Models Enable Zero-Shot Fitness Estimation}



  \icmlsetsymbol{equal}{*}

  \begin{icmlauthorlist}
    \icmlauthor{Furkan Eris}{yyy}
  \end{icmlauthorlist}

  \icmlaffiliation{yyy}{Independent researcher}

  \icmlcorrespondingauthor{Furkan Eris}{furkan6olm@gmail.com}
  \icmlkeywords{Machine Learning, ICML}

  \vskip 0.3in
]



\printAffiliationsAndNotice{}  

\begin{abstract}
Protein language models (PLMs) face a fundamental divide: masked language models (MLMs) excel at fitness prediction while causal models enable generation, forcing practitioners to maintain separate architectures. We introduce \textbf{Proust}, a 309M-parameter causal PLM that bridges this gap through architectural innovations adapted from recent LLM research, including grouped-query attention with shared K/V projections, cross-layer value residuals, and depthwise causal convolutions. Trained on 33B tokens in 40 B200 GPU-hours, Proust achieves Spearman $\rho = 0.390$ on ProteinGym substitutions, competitive with MLMs requiring 50--200$\times$ the compute. On indels, Proust sets a new state-of-the-art, outperforming models up to 20$\times$ larger. On EVEREST viral fitness benchmarks, it approaches structure-aware methods using sequence alone. These powerful representations position Proust in a sweet spot as it also retains native generative capabilities that MLMs lack by design. Interpretability analysis reveals that per-position entropy variance predicts, to an extent, when retrieval augmentation helps and hurts. Such insights can grow in both quantity and quality at scale and inform capabilities such as test-time scaling. Code and weights are available at \url{https://github.com/Furkan9015/proust-inference}
\end{abstract}

\section{Introduction}

Protein language models (PLMs) have become essential tools for predicting mutational effects, guiding directed evolution, and generating novel sequences~\cite{ProteinGym,progen2}. However, the field has split into two camps. Masked language models (MLMs) such as ESM-2~\cite{esm2} learn strong representations and score mutations via pseudo-likelihood, but cannot generate sequences directly. Causal language models (CLMs) such as ProGen2~\cite{progen2} and ProtGPT2~\cite{ProtGPT2} generate proteins autoregressively, yet have historically underperformed MLMs on fitness prediction~\cite{ProteinGym}. Practitioners needing both capabilities must maintain separate models.

We show this division is unnecessary. \textbf{Proust} is a causal PLM that reaches MLM-level fitness prediction while retaining the ability to generate sequences. On ProteinGym substitutions, Proust achieves Spearman $\rho$ within 0.01 of ESM2-650M using 62$\times$ fewer training FLOPs, and within 0.03 of E1-600M at 229$\times$ fewer FLOPs (\Cref{fig:proteingym}). On indels, Proust sets a new state-of-the-art, outperforming models up to 20$\times$ larger (\Cref{fig:indels}). On EVEREST viral fitness~\cite{EVEREST-viruses-benchmark}, it approaches structure-aware methods using sequence alone (\Cref{fig:everest}).

The efficiency gains stem from architectural choices borrowed from recent LLM research rather than protein-specific design. These include GQA-S2 attention with shared K/V projections and partial rotary embeddings~\cite{gqas2JianlinSu,RoPE}, cross-layer value residuals for improved gradient flow~\cite{valueresidual}, key offset for single-layer induction heads~\cite{moddednanogpt}, and Canon layers that add depthwise causal convolutions for local pattern recognition~\cite{CanonLayers}. Training uses the Muon optimizer~\cite{muon,NorMuon} with Newton-Schulz orthogonalization, enabling higher learning rates without instability. We detail these components in \Cref{sec:architecture}.

We also examine what Proust learns internally. Logit lens~\cite{2020logitlens} reveals how predictions form across depth. Early layers (1--6) abstract away from the input embedding, middle layers (7--16) integrate context, and late layers (17--24) converge to final predictions. The inverse logit lens shows which amino acids the model actively suppresses: tryptophan (W) at 41\% of positions and cysteine (C) at 22\%, consistent with their low frequency and specialized structural roles in natural proteins.

These internal signals turn out to be practically useful. The standard deviation of per-position entropy correlates with whether retrieval augmentation helps ($\rho = -0.40$). When uncertainty is uniform across positions (low std), external homologs provide useful signal everywhere. When uncertainty is concentrated at specific sites (high std), the model already knows which positions matter, and retrieval can overwrite correct predictions. This could offer a cheap heuristic for deciding when to run expensive homolog searches.

\paragraph{Contributions.}
\begin{itemize}[leftmargin=*,itemsep=2pt]
    \item A causal PLM that matches MLM performance on substitutions, achieves state-of-the-art among single-sequence models on indels, and retains generation capabilities, trained with 50--200$\times$ less compute than comparable models.
    \item An architecture combining GQA-S2, value residuals, key offset, and Canon layers among other optimizations and reaching 19\% MFU on B200 GPUs with 131K-token batches.
    \item Evidence that logit lens entropy statistics predict when retrieval augmentation helps, with implications for test-time compute allocation.
\end{itemize}

\section{Background}

\paragraph{Masked and causal PLMs.}
PLMs learn distributions over amino acid sequences that transfer to fitness prediction, homology detection, and structure prediction~\cite{unirep2019,prottrans,esm2,rao2019evaluating,rives2019esm}. The training objective determines what a model can do natively. MLMs such as ESM-2 and ProtTrans see bidirectional context and score mutations via pseudo-likelihood, masking position $i$, predicting $P(x_i | x_{\setminus i})$, and repeating for each position~\cite{salazar2020masked}. This requires $L$ forward passes for a length-$L$ sequence. CLMs such as ProGen2 and ProtGPT2 factorize left-to-right as $P(x) = \prod_i P(x_i | x_{<i})$. They score sequences in one pass and generate by sampling, but each position sees only leftward context.

Bidirectional context is often assumed to make MLMs better at representation tasks and more sample-efficient in limited data regimes. As we show in \Cref{sec:evaluation}, CLMs trained with modern architectures close most of this gap.

\paragraph{Retrieval augmentation.}
Single-sequence models are the simplest to deploy, requiring only the query with no database lookups or retrieval latency. However, conditioning on evolutionary context often improves accuracy. Retrieval-augmented PLMs such as PoET~\cite{PoET-2}, E1~\cite{E1-plm}, RAG-ESM~\cite{ragesm2025}, and VenusREM~\cite{VenusREM} retrieve homologs via MMseqs2~\cite{MMseqs2} or similar tools and concatenate them as context. This helps when informative homologs exist, but adds 10--60 seconds of retrieval latency per query (depending on database size), increases memory by $O(N \cdot L)$ for $N$ retrieved sequences, and risks leaking test set information if the retrieval database overlaps with evaluation sets~\cite{ProteinGym}.

Test-time scaling in LLMs trades inference compute for accuracy~\cite{cot-llm,selfconsistency,treethoughts}. In PLMs, retrieval is the dominant form of test-time scaling~\cite{testtimescalingPLMs}. We show that entropy statistics from logit lens can predict whether retrieval will help for a given protein, providing guidance on when the extra compute is justified.

\paragraph{Efficient attention with GQA-S2.}
Standard multi-head attention (MHA) stores separate K and V projections per head, with KV cache scaling as $O(n_\text{heads} \cdot d_\text{head} \cdot L)$. Grouped-query attention (GQA) shares KV heads across query groups, reducing cache by the group ratio. Prior work~\cite{gqas2JianlinSu} found that under a fixed KV cache budget, increasing head dimension matters more than increasing the number of KV groups. This motivates GQA-S2, where K and V share the same projection weights entirely, freeing parameters to increase head dimension. We split the head dimension into a no-position-encoding (NoPE) portion and a Rotary Position Embedding (RoPE) portion. Both K and V use the same projected representation, with RoPE applied to both. Since V now carries positional information, the attention output acquires an absolute position encoding. To recover relative position encoding, we apply inverse RoPE (VO-RoPE) to the RoPE portion of the output, rotating by $-\theta$ instead of $+\theta$. This configuration matches or outperforms more complex alternatives such as Multi-head Latent Attention~\cite{deepseekv2} at equivalent KV cache budgets~\cite{gqas2JianlinSu}.

\paragraph{Interpretability tools.}
Logit lens~\cite{2020logitlens} projects intermediate hidden states through the output head to reveal what the model ``would predict'' at each layer. In LLMs, predictions typically crystallize in late layers. We apply logit lens to PLMs and find similar behavior, with an added use in that entropy statistics across layers predict when retrieval augmentation will help.

Most interpretability work on PLMs has focused on sparse autoencoders, which recover features such as binding sites, secondary structure elements, and taxonomic signals~\cite{InterPLM,ProtSAE,plm-sae,plm-interp-sae}. We focus on logit lens here, though SAE features may offer complementary signals.

\section{Proust: Architecture and Training}
\label{sec:architecture}

Proust is a 309M-parameter decoder-only transformer trained with a causal language modeling objective. This section describes the architectural choices that enable competitive performance at low compute.

\subsection{Model architecture}

\paragraph{GQA-S2 attention with partial RoPE.}
We adopt GQA with the S2 scheme~\cite{gqas2JianlinSu}, where K and V share the same projection. The head dimension is split into 96 NoPE (no position encoding) dimensions and 32 RoPE dimensions. This design reflects recent findings that content matching (``what'') and position matching (``where'') play distinct roles in attention~\cite{decouplingwhatwhere,ropetonope,roundandround}. NoPE dimensions support content-based retrieval via vector similarity, while RoPE dimensions encode relative position. Mixing both in all dimensions can hurt performance when tasks require independent matching on content or position~\cite{decouplingwhatwhere}, and hybrid RoPE/NoPE strategies have accordingly outperformed pure RoPE on both short and long context tasks~\cite{ropetonope,hope-vlm}.

Both K and V come from a single linear layer, with RoPE applied to the 32-dim portion. Since V now carries positional information, the attention output has absolute position encoding. We apply VO-RoPE (i.e., inverse rotation) to this portion of the output to recover relative position encoding.

Since head dimension matters more than KV group count~\cite{gqas2JianlinSu}, we use the freed parameters from K=V sharing to increase head dimension. The configuration consists of 24 layers, hidden dim 1024, 16 query heads, 2 KV heads (8:1 ratio), and head dim 128 (96+32). We use 128 because FlashAttention-4~\cite{dao2023flashattention2} does not support head dimensions above this value at the time of writing. This yields 309M parameters and a KV cache of 512 floats per token per layer.
\paragraph{Value residuals and key offset.}
Each layer $\ell > 0$ mixes its computed values with the first layer's values~\cite{valueresidual},
\begin{equation}
V_\ell = \sigma(\lambda_\ell^{(1)}) \cdot V_\ell^\text{proj} + \sigma(\lambda_\ell^{(2)}) \cdot V_0
\end{equation}
where $\lambda_\ell^{(1,2)}$ are learned scalars initialized to $\pm 0.5$. This improves gradient flow to early layers and stabilizes representations across depth. Separately, for the NoPE dimensions of K, we shift keys forward by one position~\cite{moddednanogpt} such that $K_{\text{nope}}[t] \leftarrow K_{\text{nope}}[t-1]$. A query at position $t$ can then match against the key at $t-1$, detecting bigram patterns (e.g., ``after A comes B'') in a single layer rather than requiring the two-layer construction of standard transformers.
\vspace{-1em}
\paragraph{Canon layers.}
In causal transformers, attention handles all inter-token communication as MLPs operate position-wise. Even simple patterns like copying repeated motifs (e.g., predicting Pro after Gly in collagen's Gly-X-Y repeats, having seen Gly-Pro earlier) require two attention layers. The first copies Gly's information into its neighbor Pro, and the second retrieves Pro by matching Gly. This is expensive for patterns that could be handled with local convolutions.

Canon layers~\cite{CanonLayers} add lightweight local mixing via depthwise causal convolutions. For hidden states $h_t$,
\begin{equation}
h'_t = h_t + \text{conv1d}([h_t, h_{t-1}, h_{t-2}, h_{t-3}])
\end{equation}
where conv1d is a learned depthwise convolution with kernel size 4. The residual connection is necessary for training stability. No activation is used, as nonlinearities come from attention and FFN blocks.

We place Canon layers at three positions. Canon-A appears before attention (after pre-norm, dim = $d$), Canon-C before FFN (after pre-norm, dim = $d$), and Canon-D within FFN (after up-projection, before activation, dim = $4d$). Together with key offset, Canon-A enables single-layer induction heads for motif recognition. An additional benefit is that partial RoPE (25\%) with Canon matches or exceeds full RoPE without Canon~\cite{CanonLayers}.
\paragraph{FFN and normalization.}
We use a standard (non-gated) MLP with ReLU$^2$ activation~\cite{relusquared}, defined as $\text{FFN}(x) = W_\text{down} \cdot \text{ReLU}^2(W_\text{up} \cdot x)$ with 4$\times$ expansion. Unlike gated MLPs with SwiGLU activation~\cite{shazeer2020glu} in Llama-style models, standard MLPs with ReLU$^2$ retain more knowledge capacity in limited-exposure regimes~\cite{CanonLayers}. Canon-D is placed after up-projection but before ReLU$^2$, providing local mixing in the expanded 4$d$ space. For normalization, we use RMSNorm~\cite{RMSNorm} with sandwich normalization (pre-norm plus scaled post-norm at $1/\sqrt{L}$) and post-embedding normalization. We do not tie weights between embedding and output head.
\vspace{1em}

\subsection{Training setup}

We train on a curated dataset combining UniRef50~\cite{UniRef} version 2025\_04 with specialized databases, including OMG metagenomic proteins (coverage $>$ 3)~\cite{OMGdataset}, CAZy carbohydrate-active enzymes~\cite{CAZy}, clustered RVDB viral sequences~\cite{RVDB}, the first 10M proteins from VirE~\cite{VirE}, human-complete proteins from LOGAN~\cite{LOGAN}, BFVD~\cite{BFVD}, Arabidopsis proteins from Araport11~\cite{Araport11}, toxin sequences, and yeast strain proteins. After deduplication and filtering sequences shorter than 10 amino acids, the final dataset contains 167M sequences and 33B tokens. Sequences longer than 16,384 amino acids are randomly cropped to this maximum length. We hold out 0.04\% of sequences ($\sim$67K) for validation. The vocabulary has 21 tokens (20 amino acids plus $\langle$EOS$\rangle$) padded to 32 for hardware utilization, and we use variable-length sequence packing to avoid padding waste.

\vspace{1em}

For optimization, Muon~\cite{muon} with NorMuon improvements~\cite{NorMuon} handles attention and FFN weights using 5 iterations of Polar Express~\cite{polarexpress}, momentum 0.95, and learning rate 0.015. Polar Express computes the polar factor of the gradient, replacing it with the nearest orthogonal matrix (all singular values equal to 1), so the update has unit spectral norm ($\|\Delta W\|_* = 1$). The optimizer then rescales by $\sqrt{n_\text{out}/n_\text{in}}$. Together, these satisfy the spectral scaling condition of~\citet{yang2025spectral}, which requires $\|\Delta W\|_* = \Theta(\sqrt{n_\text{out}/n_\text{in}})$ for width-independent feature learning. This provides a theoretical basis for learning rate transfer across model widths; we verify transfer from a 50M-parameter proxy model in Appendix~\ref{app:spectral_ablation}. AdamW is used for embeddings and the output head with $\beta = (0.9, 0.95)$, learning rate $4.5 \times 10^{-4}$, and weight decay 0.01. We use a warmup-stable-decay schedule~\cite{warmupstabledecay-lrschedule} with a 10\% decay phase.

Training runs on NVIDIA B200 GPUs for 40 GPU-hours total, including distributed training overhead. Batch size is 131,072 tokens (packed with at most 768 sequences). We use FlashAttention-4~\cite{dao2023flashattention2}, torch.compile with aggressive fusion, and CUDA graphs. Model FLOPs utilization (MFU) reaches 19\%, computed as actual throughput divided by theoretical B200 peak (2.25 PFLOPs BF16). Prior to Canon layer integration, MFU was 29\%. While Canon layers add minimal parameters (less than 0.5\% for GPT2-small scale models, and as low as 0.006\% at 1.3B parameters~\cite{CanonLayers}), our current naive implementation incurs runtime overhead that reduces MFU. Fused kernel implementations could recover much of this gap, which we leave to future work. At $6.3 \times 10^{19}$ total FLOPs, Proust trains 62$\times$ faster than ESM2-650M ($3.9 \times 10^{21}$) and 229$\times$ faster than E1-600M ($1.44 \times 10^{22}$). Training completes in 62,474 steps, constituting a single pass over the dataset. Final perplexity is 10.85 on validation set and 11.04 on training set, indicating no overfitting.
\subsection{Evaluation}
\label{sec:evaluation}

\begin{figure}[h]
\centering
\includegraphics[width=\columnwidth]{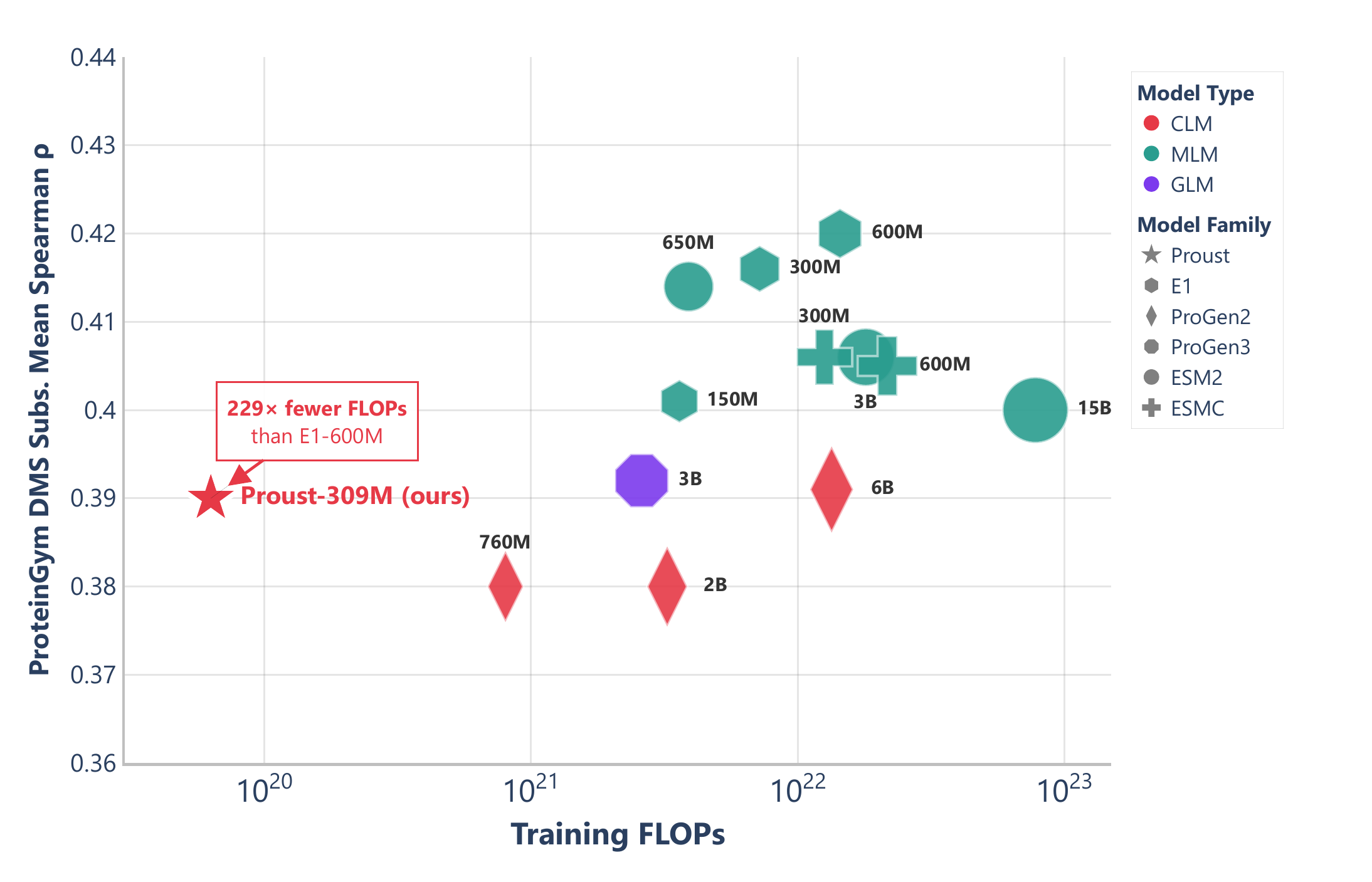}
\caption{\textbf{ProteinGym substitutions: accuracy vs.\ training compute.} Proust (star) achieves competitive Spearman $\rho$ at 229$\times$ fewer FLOPs than E1-600M. Marker shape indicates model family and color indicates training objective (CLM red, MLM teal, GLM purple). Single-sequence models are shown with solid markers and retrieval-augmented models (E1) with hollow markers.}
\label{fig:proteingym}
\end{figure}

We evaluate Proust on 217 deep mutational scanning (DMS) assays from ProteinGym~\cite{ProteinGym} against likewise single-sequence models without retrieval augmentation. For each assay, we score variants by the change in log-likelihood relative to wild-type and compute Spearman correlation with experimental fitness (\Cref{fig:proteingym}).

\paragraph{Results.}
On ProteinGym substitutions, Proust achieves $\rho = 0.390$, matching ProGen2-6.4B ($\rho = 0.391$) and ProGen3-3B ($\rho = 0.392$)~\cite{progen3} while using 41--213$\times$ less training compute. ESM2-650M reaches $\rho = 0.414$ at 62$\times$ the compute, and E1-600M reaches $\rho = 0.420$ at 229$\times$ the compute. Proust matches or exceeds all CLM and GLM models; higher correlation requires MLM objectives with substantially more resources.

For baseline models, we report numbers from the respective papers and include only models that reported single-sequence performance. Retrieval-augmented models (E1, PoET, VenusREM) incur additional inference-time compute for homolog search not reflected in training FLOPs, and \Cref{fig:proteingym} distinguishes single-sequence from retrieval-augmented models. Breaking down by functional category, Proust performs best on activity assays ($\rho = 0.42$) and binding assays ($\rho = 0.41$), where sequence-level patterns often correlate with function. Performance is lower on stability assays ($\rho = 0.34$), which depend more heavily on 3D structural context that sequence models capture only indirectly. This pattern is consistent across PLMs, as stability prediction remains challenging without explicit structural information.

\begin{figure}[h]
\centering
\includegraphics[width=\columnwidth]{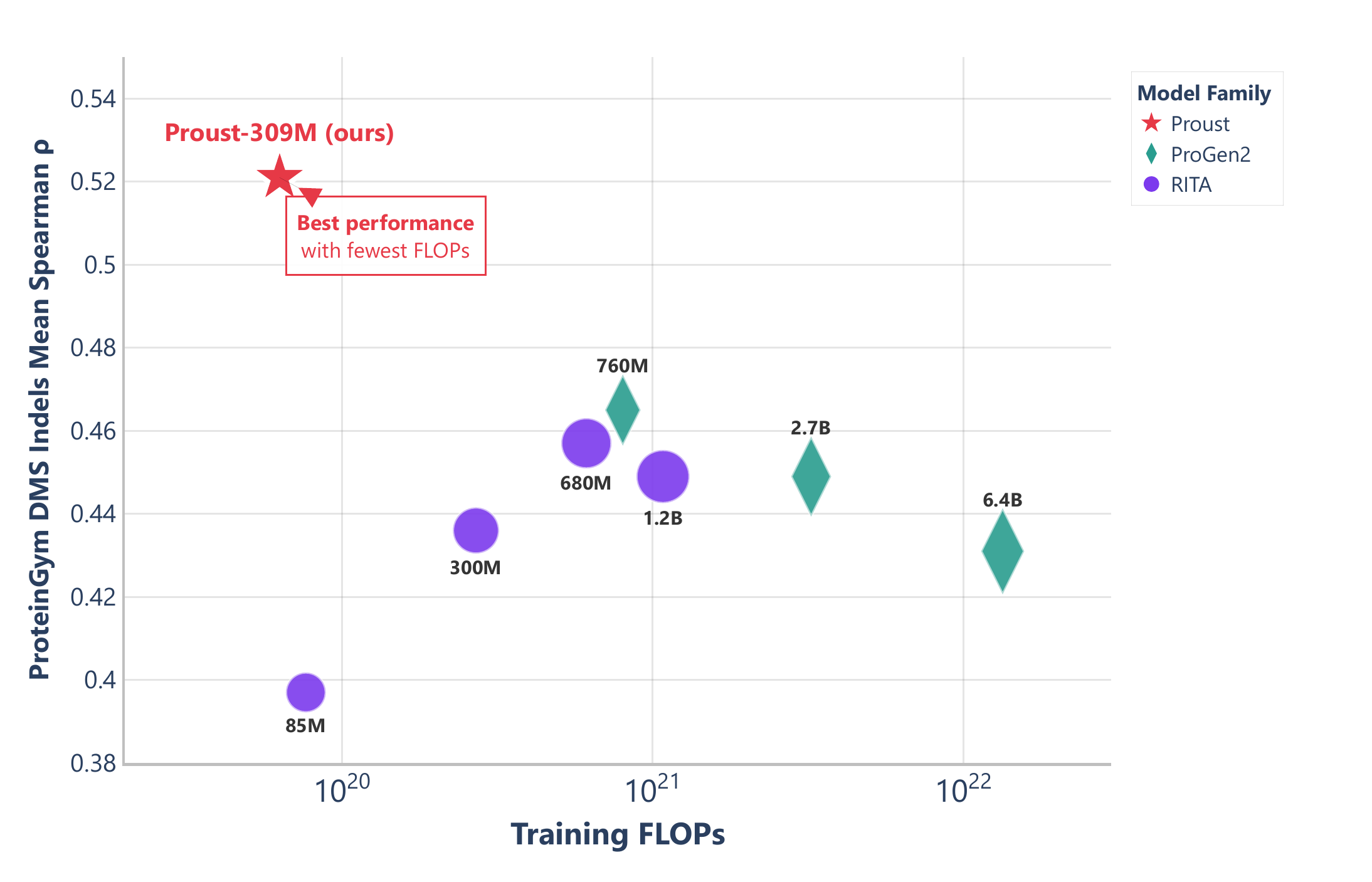}
\caption{\textbf{ProteinGym indels: accuracy vs.\ training compute.} Proust achieves the highest Spearman $\rho$ among all compared models while using the fewest training FLOPs. Causal models handle indels naturally since insertions and deletions do not disrupt autoregressive scoring.}
\label{fig:indels}
\end{figure}

On ProteinGym indels (69 assays), Proust achieves $\rho = 0.521$, outperforming all compared models including ProGen2-6.4B ($\rho = 0.432$) and RITA-1.2B~\cite{RITA} ($\rho = 0.450$) despite using 10--200$\times$ less training compute (\Cref{fig:indels}). Indels are a natural fit for causal models because insertions and deletions shift all downstream positions, breaking the fixed-position assumptions of MLMs but posing no difficulty for autoregressive scoring. The gap between Proust and larger models is larger for indels than for substitutions, suggesting that architectural efficiency matters more than raw parameter count for this task.

\begin{figure}[h]
\centering
\includegraphics[width=\columnwidth]{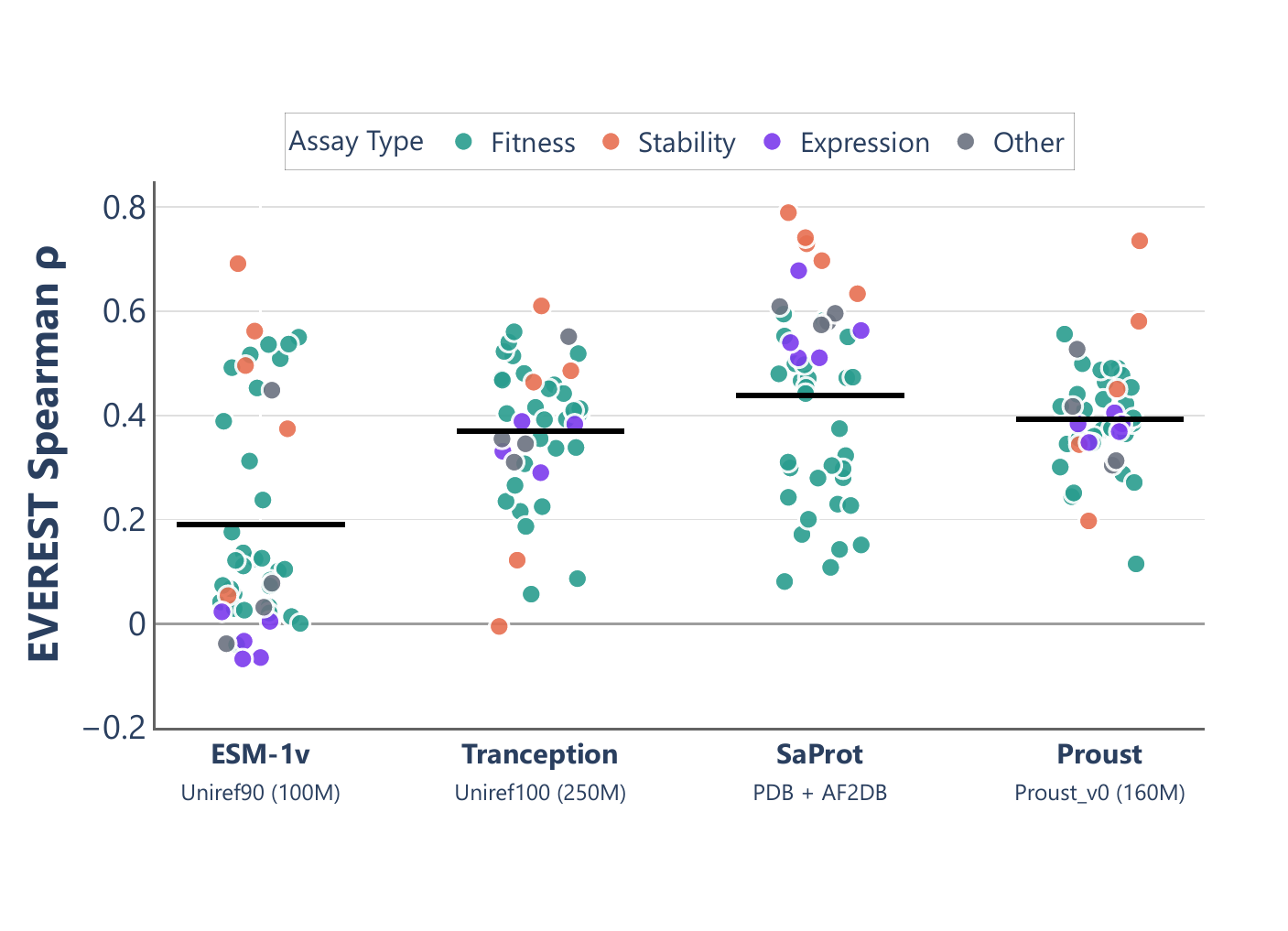}
\caption{\textbf{EVEREST viral fitness benchmark.} Points are colored by assay type and horizontal bars indicate mean $\rho$ per model. Proust shows lower variance across assays than baselines, avoiding failure modes but missing peaks on stability assays where structure-aware models (SaProt) excel.}
\label{fig:everest}
\end{figure}

We also evaluate on EVEREST~\cite{EVEREST-viruses-benchmark}, which contains 45 viral DMS assays (\Cref{fig:everest}). Proust achieves mean $\rho = 0.40$, approaching SaProt~\cite{SaProt} ($\rho = 0.44$) which uses structure tokens from Foldseek~\cite{foldseek}. Viral proteins present a distinct challenge because they evolve rapidly, have fewer homologs in standard databases, and often contain disordered regions.

Notably, Proust exhibits lower cross-assay variance than other models ($\sigma = 0.10$ vs.\ $0.14$--$0.22$ for baselines). This suggests more consistent performance, as Proust avoids the failure modes that cause ESM-1v~\cite{esm1v} and SaProt-AF2 to achieve near-zero or negative correlations on some assays, but also misses the peaks where structure-aware models excel. The largest gap appears on stability assays, where SaProt-PDB reaches $\rho > 0.70$ while Proust remains near $\rho = 0.45$. This is consistent with prior work showing that thermostability prediction benefits from explicit structural information~\cite{ProSST}.

\section{Test-time Scaling: When Does Retrieval Help?}

Retrieval augmentation (conditioning on homologs at inference time) improves accuracy on many fitness prediction tasks but adds cost. Retrieving homologs via ColabFold~\cite{colabfold} to build a multiple sequence alignment (MSA) takes 10--60 seconds per query, depending on sequence length, server load, and reference database (see Appendix~\ref{app:retrieval_latency}). For the 217 ProteinGym substitution assays, retrieval adds approximately 4 hours of wall-clock time when run sequentially.

\subsection{PSSM-based scoring from homologs}

We retrieve MSAs via ColabFold API using UniRef30 and environmental databases (BFD~\cite{bfd}, MGnify~\cite{mgnify}, MetaEuk~\cite{metaeuk}). We filter out homologs with $\leq$50\% coverage of the target sequence to exclude fragments and select the $N$ most similar homologs by sequence identity. From these homologs, we construct position-specific scoring matrices (PSSMs)~\cite{Gribskov1987}, where each entry is a log-odds score against a uniform background: $\text{PSSM}[i, a] = \log_2(f_{i,a} / 0.05)$, with $f_{i,a}$ the observed frequency of amino acid $a$ at position $i$.

To score a variant, we sum PSSM contributions only at mutated positions:
\begin{equation}
S_{\text{PSSM}} = \sum_{i \in \mathcal{M}} \big(\text{PSSM}[i, a_i^{\text{mut}}] - \text{PSSM}[i, a_i^{\text{wt}}]\big)
\end{equation}
where $\mathcal{M}$ is the set of positions that differ between variant and wild-type. This focuses evolutionary signal on the sites under evaluation rather than diluting it across the full sequence.

The final combined score averages z-normalized model log-likelihood and PSSM scores with equal weight:
\begin{equation}
S_{\text{combined}} = \frac{1}{2}\hat{S}_{\text{LL}} + \frac{1}{2}\hat{S}_{\text{PSSM}}
\end{equation}
where $\hat{S} = (S - \mu_S)/\sigma_S$ denotes z-normalization across variants within each assay. This ensures both components contribute equally regardless of their raw scale.

\begin{figure}[h]
\centering
\includegraphics[width=\columnwidth]{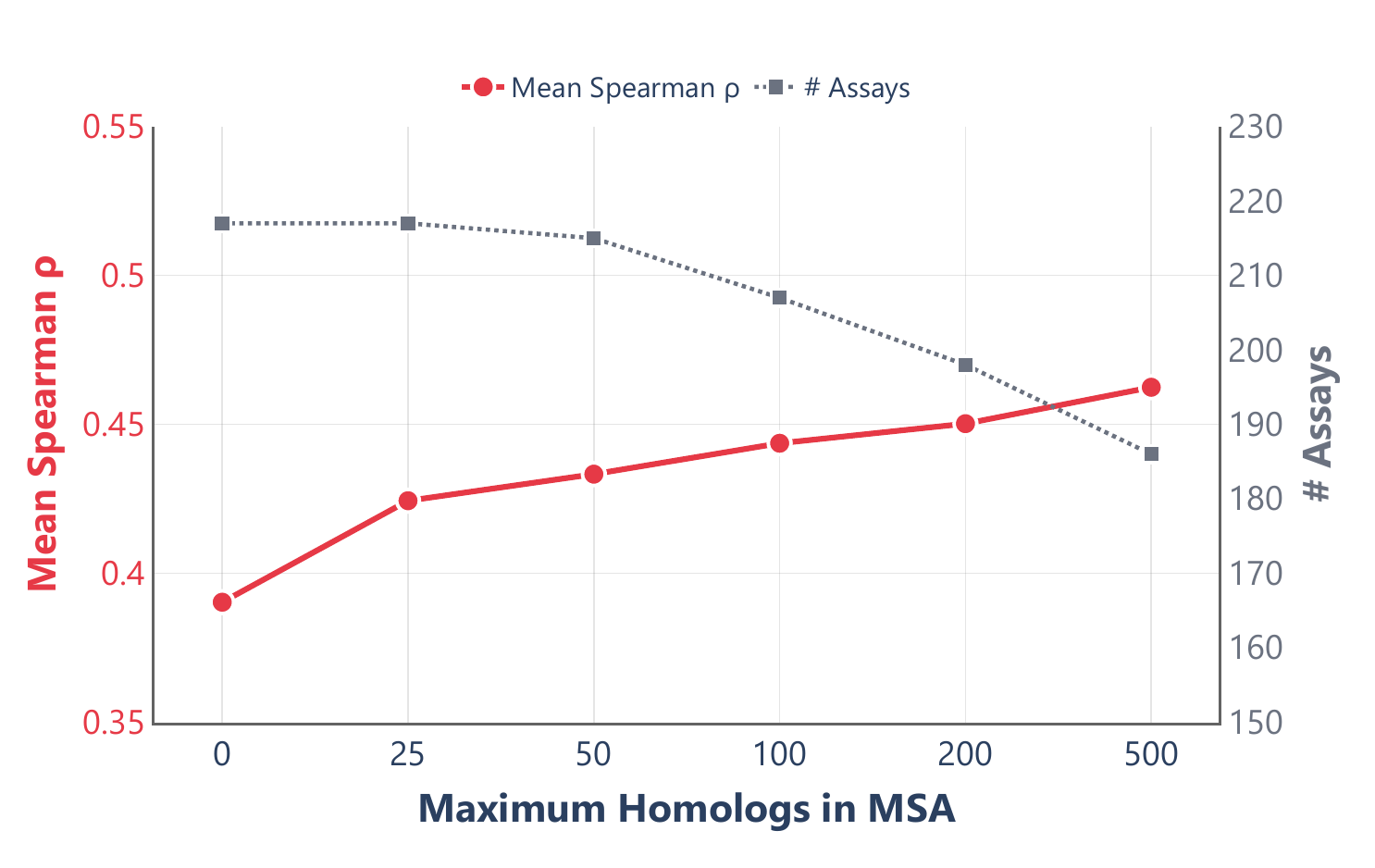}
\caption{\textbf{Effect of homolog depth on ProteinGym substitution performance.} We select the $N$ most similar homologs (by sequence identity, $\geq$50\% coverage). Mean Spearman $\rho$ (red, left axis) improves with depth. The number of assays with sufficient homologs (gray, right axis) decreases at higher depths, as not all proteins have deep MSAs.}
\label{fig:homolog_depth}
\end{figure}

\Cref{fig:homolog_depth} shows the effect of homolog depth on PSSM-augmented scoring for ProteinGym substitutions. Mean Spearman $\rho$ increases monotonically with depth, from 0.390 (no retrieval) to 0.432 at depth 500. However, not all assays have deep MSAs, and the number of qualifying assays decreases from 217 at depth 0 to $\sim$170 at depth 500. For proteins with sufficient evolutionary coverage, PSSM augmentation provides consistent gains.

\section{Mechanistic Interpretation of Proust}
\label{sec:interpretation}

Beyond fitness prediction, we wanted to understand what Proust actually learns. We apply logit lens and attention analysis to examine the model's internal representations, looking for biological knowledge that emerges from training and for signals that might predict when retrieval helps.

\paragraph{Probing transformer internals.}
The logit lens~\citep{nostalgebraist2020logitlens} projects each layer's hidden states into vocabulary space via the unembedding matrix, showing how predictions form with depth. Representational drift between layers can make these projections unreliable, so the tuned lens~\citep{belrose2023tuned} trains affine probes per layer to improve consistency. Patchscopes~\citep{ghandeharioun2024patchscopes} unifies these projection-based methods by patching hidden states between prompts. Separately, probing classifiers~\citep{alain2017probing} train linear predictors on frozen representations to test what information is linearly accessible, and activation patching~\citep{meng2022locating} measures causal effects of specific activations through targeted interventions. We focus on logit lens and attention analysis here, as they require no additional training and apply directly to Proust.

We organize our findings into three themes: (1) prediction dynamics across layers, (2) learned biological constraints, and (3) emergent structural awareness.

\paragraph{Layer-wise prediction dynamics.}
Logit lens applied to Proust shows a U-shaped accuracy curve across 24 layers. Layers 1--6 have near-zero prediction accuracy even though they receive token embeddings directly, suggesting the model discards raw token identity early on. Layers 7--16 gradually recover predictive signal. Layers 17--24 show rapid convergence to final predictions. Text language models behave similarly: early layers build abstract representations before later layers reconstruct outputs~\citep{belrose2023tuned}. This pattern suggests Proust computes contextual features in early/middle layers before committing to amino acid predictions in late layers.

\subsection{Amino acid prediction biases}

Proust over-predicts common amino acids, including serine (15.4\% predicted vs 7.1\% in data, 2.2$\times$) and leucine (14.6\% vs 9.0\%, 1.6$\times$). It under-predicts glutamine (2.4$\times$ below frequency) and methionine (2.1$\times$ below). Using the inverse logit lens (negating hidden states before projection), we find the model actively suppresses tryptophan at 41\% of positions and cysteine at 22\%. This matches biology, as tryptophan is the most energetically costly amino acid to synthesize~\cite{Akashi2002}, and cysteine residues forming disulfide bonds are under strong evolutionary constraint~\cite{Wong2011}.
\vspace{-1em}
\subsection{Positional patterns}

Entropy varies systematically with sequence position (\Cref{tab:position}). N- and C-termini show higher entropy, while the core region (40--70\% of sequence) shows the lowest. This matches evolutionary data: analysis of primate and plant gene families found elevated non-synonymous mutation rates ($d_N$) in the first and last $\sim$50 codons, producing a U-shaped pattern along the coding sequence~\cite{Bricout2023}. Termini are enriched in intrinsically disordered regions~\cite{Lobley2007} and often contain signal peptides, which tolerate sequence variability so long as their hydrophobic and polar regions are preserved~\cite{Bagos2008}. Protein cores, by contrast, are under strong negative selection. Circular permutation experiments show that relocating termini into fast-folding cores disrupts both function and solubility~\cite{Dall2024}. The model appears to have learned this constraint from sequence statistics alone: positions that are variable in nature receive higher entropy predictions.

\begin{table}[h]
\centering
\small
\begin{tabular}{@{}lcc@{}}
\toprule
Region & Relative Position & Mean Entropy \\
\midrule
N-terminus & 0--10\% & 2.07 \\
N-terminal & 10--20\% & 1.75 \\
Core & 40--70\% & 1.42--1.54 \\
C-terminal & 80--90\% & 1.53 \\
C-terminus & 90--100\% & 1.74 \\
\bottomrule
\end{tabular}
\caption{\textbf{Positional entropy patterns.} The model is most uncertain at termini, most confident in the structural core.}
\label{tab:position}
\end{table}
\vspace{-2em}
\subsection{Attention patterns}

Averaging attention weights across ProteinGym DMS Substitutions protein sequences yields a surprising finding. 65\% of attention weight goes to positions $>$20 residues away, while only 29\% goes to local positions ($\leq$10 residues). Local patterns (e.g., bigrams, trigrams, short-range dependencies) should be easier to learn, yet the model allocates most capacity to distant positions. This long-range bias suggests the model captures distal contacts, meaning residue pairs that are far apart in sequence but close in 3D structure due to protein folding. Such contacts are a hallmark of tertiary structure and cannot be predicted from local sequence alone.

Attention also correlates with amino acid properties. Hydrophobic residues (LAVIMFW) receive the highest mean attention weight (0.0106), followed by charged (DEKR, 0.0102), polar (STNQYH, 0.0092), and special residues (GPC, 0.0091). The elevated attention to hydrophobic residues is consistent with hydrophobic core packing being a dominant constraint in globular proteins~\cite{Richards1977}, suggesting the model tracks which positions participate in the buried core. The lower attention to glycine, proline, and cysteine may reflect their specialized roles (backbone flexibility, helix breaking, and disulfide bonding, respectively), which depend more on local context than long-range interactions.

\subsection{Biophysical understanding}

Proust captures hydrophobic clustering without explicit supervision. The relationship between local context and prediction is monotonic. As the fraction of hydrophobic residues in the surrounding 5-residue window increases, so does the probability that the model predicts a hydrophobic amino acid (from 20\% in hydrophilic contexts to 61\% in strongly hydrophobic contexts). The correlation between local hydrophobic fraction and predicted hydrophobic probability is $\rho = 0.32$ (p $< 10^{-100}$). This gradient mirrors the physical tendency of hydrophobic residues to cluster in protein cores, where they minimize contact with solvent.

Secondary structure propensities also emerge from the training objective. Helix-favoring amino acids (A, E, L, M) show $\rho = 0.33$ correlation between local helix context and prediction probability, while sheet-favoring residues (V, I, Y, W) show $\rho = 0.20$. The stronger signal for helices likely reflects their more regular sequence patterns (i, i+3, i+4 spacing) compared to the irregular hydrogen-bonding patterns in beta sheets~\cite{Kumar1998}.

\subsection{Motif recognition}

We test whether Proust recognizes functional sequence motifs by comparing entropy at motif positions versus non-motif positions (\Cref{tab:motifs}). If the model has learned that certain positions are functionally constrained, it should show lower entropy (higher confidence) at those positions.

\begin{table}[h]
\centering
\small
\begin{tabular}{@{}llcc@{}}
\toprule
Motif & Function & Entropy Ratio & Interpretation \\
\midrule
CxxC & Zinc finger & 0.60 & Highly constrained \\
NxS/T & N-glycosylation & 1.04 & Recognized \\
GxxG & P-loop & 1.05 & Slightly constrained \\
PxxP & SH3 binding & 1.12 & Variable \\
\bottomrule
\end{tabular}
\caption{\textbf{Motif entropy analysis.} Ratio $<$1 indicates the model treats motif positions as more constrained than background. CxxC shows 40\% lower entropy, indicating strong recognition of this zinc-finger motif.}
\label{tab:motifs}
\end{table}

The CxxC zinc-finger motif shows the strongest signal, with 40\% lower entropy than non-motif positions (1.55 vs 2.60). Zinc fingers require precise cysteine spacing to coordinate metal ions, and mutations at these positions typically abolish function~\cite{Harding2004,Eom2016,Godwin2017}. The low entropy suggests the model has internalized this constraint purely from sequence co-occurrence patterns. The N-glycosylation sequon (NxS/T) and P-loop (GxxG) motifs show entropy ratios near 1.0, indicating the model recognizes these patterns but treats the variable positions (x) as rather flexible. The PxxP motif, which mediates SH3 domain binding, shows slightly elevated entropy (ratio 1.12), in line with the known tolerance of this motif to substitutions at the proline-flanking positions~\cite{Nguyen2000,Li2005,Gorelik2012}.

\section{Conclusion}

Proust is a 309M-parameter causal protein language model trained at 50--200$\times$ lower compute cost than ESM2-650M~\cite{esm2} and E1-600M~\cite{E1-plm}. On ProteinGym~\cite{ProteinGym} substitutions, it matches MLM-based models that required an order of magnitude more training FLOPs. On indels, it outperforms all compared models including those 20$\times$ its size. Because Proust is autoregressive, it retains the ability to generate sequences, a capability that MLMs lack by design.

The efficiency comes from architectural choices developed for text language models, including GQA-S2 attention with shared KV projections~\cite{gqas2JianlinSu}, cross-layer value residuals~\cite{valueresidual}, key offset for induction heads~\cite{moddednanogpt}, Canon layers for local patterns~\cite{CanonLayers}, and the Muon optimizer for stable high learning rates~\cite{muon}. None of these are protein-specific, which suggests they may transfer to other modalities where autoregressive modeling is natural.

Applying interpretability methods from language modeling, we find that Proust has learned to distinguish constrained positions from variable ones without explicit structural supervision. Prediction entropy is lowest in protein cores and highest at termini, matching known patterns of evolutionary constraint. Attention concentrates on distant residues rather than local ones, consistent with the role of tertiary contacts in determining sequence. The model suppresses tryptophan and cysteine at most positions, reflecting their rarity and specialized structural roles. Functional motifs like CxxC zinc fingers show sharply reduced entropy, indicating the model has internalized their conservation.

These internal signals have practical value. The standard deviation of per-position entropy predicts whether retrieval augmentation will help for a given protein ($\rho = -0.40$). When uncertainty is uniform, homologs provide useful signals. When concentrated at specific sites, the model already knows what matters and retrieval can hurt. This holds the potential to offer a cheap heuristic for allocating test-time compute.

Proust does have limitations. Stability prediction lags behind structure-aware models like SaProt~\cite{SaProt}, consistent with the difficulty of capturing thermodynamic properties from sequence alone. Retrieval-augmented MLMs still achieve higher absolute correlations when compute is not a constraint, and whether scaling Proust to 1--3B parameters closes these gaps remains an open question.

\section{Future Work}

Our results suggest several directions worth exploring.

\paragraph{Training dataset.} We use single sequences during pretraining but concatenate homologs only at test time. Pretraining on concatenated homologs (as in PoET~\cite{PoET-1} and E1~\cite{E1-plm}) might improve retrieval-augmented performance. This is especially feasible given Proust's existing long context pretraining, though it requires batching strategies tailored to this setting. Next, same prior works attempt a balanced sampling out of protein families with different sizes. However, this requires a precomputed all-vs-all alignment of training data which becomes increasingly overwhelming as dataset size grows. This underscores the potential of lightweight subset selection methods from the LLM literature towards improving the data aspect of PLM pretraining.

\paragraph{Scaling.} Proust is deliberately small (309M) to demonstrate efficiency. Scaling to 1--3B parameters with the same architecture should improve performance at equal Chinchillas of data while maintaining the FLOP advantage over prior work.

\paragraph{Muon warmup.} We use zero warmup for the Muon optimizer based on preliminary experiments, but systematic study of warmup schedules for Newton-Schulz-based optimizers remains open.

\paragraph{Homolog etrieval databases.} Our retrieval uses ColabFold's environmental mode, which searches UniRef30 and metagenomic databases (BFD~\cite{bfd}, MGnify~\cite{mgnify}, MetaEuk~\cite{metaeuk}). Larger databases such as UniRef100 or the OMG dataset~\cite{OMGdataset} could improve coverage for proteins with low MSA depth, but would increase retrieval latency.

\section*{Impact Statement}

This paper presents a protein language model for fitness prediction and sequence generation. Improved fitness prediction could accelerate protein engineering for therapeutic and industrial applications. We note that Proust operates on sequence alone and does not predict or optimize for specific functions; capable models with similar functionality are already publicly available. Model weights will be released to enable reproducibility.


\bibliography{proust_paper}

@article{Gorelik2012,
  title = {Distinct Peptide Binding Specificities of Src Homology 3 (SH3) Protein Domains Can Be Determined by Modulation of Local Energetics across the Binding Interface},
  volume = {287},
  ISSN = {0021-9258},
  url = {http://dx.doi.org/10.1074/jbc.m111.330753},
  DOI = {10.1074/jbc.m111.330753},
  number = {12},
  journal = {Journal of Biological Chemistry},
  publisher = {Elsevier BV},
  author = {Gorelik, Maryna and Davidson, Alan R.},
  year = {2012},
  month = mar,
  pages = {9168--9177}
}

@article{Li2005,
  title = {Specificity and versatility of SH3 and other proline-recognition domains: structural basis and implications for cellular signal transduction},
  volume = {390},
  ISSN = {1470-8728},
  url = {http://dx.doi.org/10.1042/BJ20050411},
  DOI = {10.1042/bj20050411},
  number = {3},
  journal = {Biochemical Journal},
  publisher = {Portland Press Ltd.},
  author = {Li, Shawn S.-C.},
  year = {2005},
  month = sep,
  pages = {641--653}
}

@article{Nguyen2000,
  title = {Improving SH3 domain ligand selectivity using a non-natural scaffold},
  volume = {7},
  ISSN = {1074-5521},
  url = {http://dx.doi.org/10.1016/s1074-5521(00)00130-7},
  DOI = {10.1016/s1074-5521(00)00130-7},
  number = {7},
  journal = {Chemistry \& Biology},
  publisher = {Elsevier BV},
  author = {Nguyen, Jack T and Porter, Margaret and Amoui, Mehran and Miller, W Todd and Zuckermann, Ronald N and Lim, Wendell A},
  year = {2000},
  month = jul,
  pages = {463--473}
}

@article{Eom2016,
  title = {Structural Analyses of Zinc Finger Domains for Specific Interactions with DNA},
  volume = {26},
  ISSN = {1738-8872},
  url = {http://dx.doi.org/10.4014/jmb.1609.09021},
  DOI = {10.4014/jmb.1609.09021},
  number = {12},
  journal = {Journal of Microbiology and Biotechnology},
  publisher = {Korean Society for Microbiology and Biotechnology},
  author = {Eom, Ki Seong and Cheong, Jin Sung and Lee, Seung Jae},
  year = {2016},
  month = dec,
  pages = {2019--2029}
}

@article{Godwin2017,
  title = {All-atom molecular dynamics comparison of disease-associated zinc fingers},
  volume = {36},
  ISSN = {1538-0254},
  url = {http://dx.doi.org/10.1080/07391102.2017.1363662},
  DOI = {10.1080/07391102.2017.1363662},
  number = {10},
  journal = {Journal of Biomolecular Structure and Dynamics},
  publisher = {Informa UK Limited},
  author = {Godwin, Ryan C. and Gmeiner, William H. and Salsbury, Freddie R.},
  year = {2017},
  month = oct,
  pages = {2581--2594}
}

@article{Harding2004,
  title = {The architecture of metal coordination groups in proteins},
  volume = {60},
  ISSN = {0907-4449},
  url = {http://dx.doi.org/10.1107/S0907444904004081},
  DOI = {10.1107/s0907444904004081},
  number = {5},
  journal = {Acta Crystallographica Section D Biological Crystallography},
  publisher = {International Union of Crystallography (IUCr)},
  author = {Harding, Marjorie M.},
  year = {2004},
  month = apr,
  pages = {849--859}
}

@article{Richards1977,
  title = {Areas, Volumes, Packing, and Protein Structure},
  volume = {6},
  ISSN = {0084-6589},
  url = {http://dx.doi.org/10.1146/annurev.bb.06.060177.001055},
  DOI = {10.1146/annurev.bb.06.060177.001055},
  number = {1},
  journal = {Annual Review of Biophysics and Bioengineering},
  publisher = {Annual Reviews},
  author = {Richards, Frederic M.},
  year = {1977},
  month = jun,
  pages = {151--176}
}

@article{Kumar1998,
  title = {Geometrical and Sequence Characteristics of $\alpha$-Helices in Globular Proteins},
  volume = {75},
  ISSN = {0006-3495},
  url = {http://dx.doi.org/10.1016/S0006-3495(98)77634-9},
  DOI = {10.1016/s0006-3495(98)77634-9},
  number = {4},
  journal = {Biophysical Journal},
  publisher = {Elsevier BV},
  author = {Kumar, Sandeep and Bansal, Manju},
  year = {1998},
  month = oct,
  pages = {1935--1944}
}

@article{Akashi2002,
  title = {Metabolic efficiency and amino acid composition in the proteomes of {Escherichia coli} and {Bacillus subtilis}},
  volume = {99},
  ISSN = {1091-6490},
  url = {http://dx.doi.org/10.1073/pnas.062526999},
  DOI = {10.1073/pnas.062526999},
  number = {6},
  journal = {Proceedings of the National Academy of Sciences},
  publisher = {National Academy of Sciences},
  author = {Akashi, Hiroshi and Gojobori, Takashi},
  year = {2002},
  month = mar,
  pages = {3695--3700}
}

@article{Wong2011,
  title = {Disulfide Bond Acquisition through Eukaryotic Protein Evolution},
  volume = {28},
  ISSN = {1537-1719},
  url = {http://dx.doi.org/10.1093/molbev/msq194},
  DOI = {10.1093/molbev/msq194},
  number = {1},
  journal = {Molecular Biology and Evolution},
  publisher = {Oxford University Press (OUP)},
  author = {Wong, Jason W. H. and Ho, Simon Y. W. and Hogg, Philip J.},
  year = {2011},
  month = jan,
  pages = {327--334}
}

@article{Dall2024,
  title = {The importance of the location of the N-terminus in successful protein folding in vivo and in vitro},
  volume = {121},
  ISSN = {1091-6490},
  url = {http://dx.doi.org/10.1073/pnas.2321999121},
  DOI = {10.1073/pnas.2321999121},
  number = {34},
  journal = {Proceedings of the National Academy of Sciences},
  publisher = {Proceedings of the National Academy of Sciences},
  author = {Dall,  Natalie R. and Mendon\c{c}a,  Carolina A. T. F. and Torres Vera,  Héctor L. and Marqusee,  Susan},
  year = {2024},
  month = aug
}

@article{Bricout2023,
  title = {Evolution is not Uniform Along Coding Sequences},
  volume = {40},
  ISSN = {1537-1719},
  url = {http://dx.doi.org/10.1093/molbev/msad042},
  DOI = {10.1093/molbev/msad042},
  number = {3},
  journal = {Molecular Biology and Evolution},
  publisher = {Oxford University Press (OUP)},
  author = {Bricout,  Raphaël and Weil,  Dominique and Stroebel,  David and Genovesio,  Auguste and Roest Crollius,  Hugues},
  editor = {Echave,  Julian},
  year = {2023},
  month = feb
}

@article{Bagos2008,
  title = {Prediction of signal peptides in archaea},
  volume = {22},
  ISSN = {1741-0134},
  url = {http://dx.doi.org/10.1093/protein/gzn064},
  DOI = {10.1093/protein/gzn064},
  number = {1},
  journal = {Protein Engineering Design and Selection},
  publisher = {Oxford University Press (OUP)},
  author = {Bagos,  P.G. and Tsirigos,  K.D. and Plessas,  S.K. and Liakopoulos,  T.D. and Hamodrakas,  S.J.},
  year = {2008},
  month = nov,
  pages = {27–35}
}

@article{Lobley2007,
  title = {Inferring Function Using Patterns of Native Disorder in Proteins},
  volume = {3},
  ISSN = {1553-7358},
  url = {http://dx.doi.org/10.1371/journal.pcbi.0030162},
  DOI = {10.1371/journal.pcbi.0030162},
  number = {8},
  journal = {PLoS Computational Biology},
  publisher = {Public Library of Science (PLoS)},
  author = {Lobley,  Anna and Swindells,  Mark B and Orengo,  Christine A and Jones,  David T},
  editor = {Rost,  Burkhard},
  year = {2007},
  month = aug,
  pages = {e162}
}

@misc{nostalgebraist2020logitlens,
  author = {nostalgebraist},
  title = {interpreting {GPT}: the logit lens},
  year = {2020},
  howpublished = {LessWrong},
  url = {https://www.lesswrong.com/posts/AcKRB8wDpdaN6v6ru/interpreting-gpt-the-logit-lens}
}

@inproceedings{belrose2023tuned,
  title = {Eliciting Latent Predictions from Transformers with the Tuned Lens},
  author = {Belrose, Nora and Furman, Zach and Smith, Logan and Halawi, Danny and McKinney, Lev and Ostrovsky, Igor and Biderman, Stella and Steinhardt, Jacob},
  booktitle = {Advances in Neural Information Processing Systems},
  year = {2023},
  url = {https://arxiv.org/abs/2303.08112}
}

@inproceedings{ghandeharioun2024patchscopes,
  title = {Patchscopes: A Unifying Framework for Inspecting Hidden Representations of Language Models},
  author = {Ghandeharioun, Asma and Caciularu, Avi and Pearce, Adam and Dixon, Lucas and Geva, Mor},
  booktitle = {Proceedings of the 41st International Conference on Machine Learning},
  year = {2024},
  url = {https://arxiv.org/abs/2401.06102}
}

@inproceedings{alain2017probing,
  title = {Understanding intermediate layers using linear classifier probes},
  author = {Alain, Guillaume and Bengio, Yoshua},
  booktitle = {5th International Conference on Learning Representations, ICLR 2017, Workshop Track Proceedings},
  year = {2017},
  url = {https://arxiv.org/abs/1610.01644}
}

@inproceedings{meng2022locating,
  title = {Locating and Editing Factual Associations in {GPT}},
  author = {Meng, Kevin and Bau, David and Andonian, Alex and Belinkov, Yonatan},
  booktitle = {Advances in Neural Information Processing Systems},
  volume = {35},
  pages = {17359--17372},
  year = {2022},
  url = {https://arxiv.org/abs/2202.05262}
}

@misc{RITA,
  doi = {10.48550/ARXIV.2205.05789},
  url = {https://arxiv.org/abs/2205.05789},
  author = {Hesslow, Daniel and Zanichelli, Niccol\'{o} and Notin, Pascal and Poli, Iacopo and Marks, Debora},
  title = {{RITA}: a Study on Scaling Up Generative Protein Sequence Models},
  publisher = {arXiv},
  year = {2022}
}

@article{esm1v,
  title = {Language models enable zero-shot prediction of the effects of mutations on protein function},
  volume = {35},
  journal = {Advances in Neural Information Processing Systems},
  author = {Meier, Joshua and Rao, Roshan and Verkuil, Robert and Liu, Jason and Sercu, Tom and Rives, Alexander},
  year = {2021},
  pages = {29287--29303}
}

@article{esm2,
  title = {Evolutionary-scale prediction of atomic-level protein structure with a language model},
  volume = {379},
  ISSN = {1095-9203},
  url = {http://dx.doi.org/10.1126/science.ade2574},
  DOI = {10.1126/science.ade2574},
  number = {6637},
  journal = {Science},
  publisher = {American Association for the Advancement of Science (AAAS)},
  author = {Lin,  Zeming and Akin,  Halil and Rao,  Roshan and Hie,  Brian and Zhu,  Zhongkai and Lu,  Wenting and Smetanin,  Nikita and Verkuil,  Robert and Kabeli,  Ori and Shmueli,  Yaniv and dos Santos Costa,  Allan and Fazel-Zarandi,  Maryam and Sercu,  Tom and Candido,  Salvatore and Rives,  Alexander},
  year = {2023},
  month = mar,
  pages = {1123–1130}
}

@article{progen3,
  title = {Scaling Unlocks Broader Generation and Deeper Functional Understanding of Proteins},
  url = {http://dx.doi.org/10.1101/2025.04.15.649055},
  DOI = {10.1101/2025.04.15.649055},
  publisher = {openRxiv},
  author = {Bhatnagar,  Aadyot and Jain,  Sarthak and Beazer,  Joel and Curran,  Samuel C. and Hoffnagle,  Alexander M. and Ching,  Kyle S. and Martyn,  Michael and Nayfach,  Stephen and Ruffolo,  Jeffrey A. and Madani,  Ali},
  year = {2025},
  month = apr 
}

@article{PoET-1,
  doi = {10.48550/ARXIV.2306.06156},
  url = {https://arxiv.org/abs/2306.06156},
  author = {Truong,  Timothy F. and Bepler,  Tristan},
  title = {PoET: A generative model of protein families as sequences-of-sequences},
  publisher = {arXiv},
  year = {2023},
  copyright = {arXiv.org perpetual,  non-exclusive license}
}

@misc{PoET-2,
  doi = {10.48550/ARXIV.2508.04724},
  url = {https://arxiv.org/abs/2508.04724},
  author = {Truong,  Timothy Fei and Bepler,  Tristan},
  title = {Understanding protein function with a multimodal retrieval-augmented foundation model},
  publisher = {arXiv},
  year = {2025},
  copyright = {arXiv.org perpetual,  non-exclusive license}
}

@article{E1-plm,
  title = {E1: Retrieval-Augmented Protein Encoder Models},
  url = {http://dx.doi.org/10.1101/2025.11.12.688125},
  DOI = {10.1101/2025.11.12.688125},
  publisher = {openRxiv},
  author = {Jain,  Sarthak and Beazer,  Joel and Ruffolo,  Jeffrey A. and Bhatnagar,  Aadyot and Madani,  Ali},
  year = {2025},
  month = nov 
}

@inproceedings{ProteinGym,
 author = {Notin, Pascal and Kollasch, Aaron and Ritter, Daniel and van Niekerk, Lood and Paul, Steffanie and Spinner, Han and Rollins, Nathan and Shaw, Ada and Orenbuch, Rose and Weitzman, Ruben and Frazer, Jonathan and Dias, Mafalda and Franceschi, Dinko and Gal, Yarin and Marks, Debora},
 booktitle = {Advances in Neural Information Processing Systems},
 editor = {A. Oh and T. Naumann and A. Globerson and K. Saenko and M. Hardt and S. Levine},
 pages = {64331--64379},
 publisher = {Curran Associates, Inc.},
 title = {ProteinGym: Large-Scale Benchmarks for Protein Fitness Prediction and Design},
 url = {https://proceedings.neurips.cc/paper_files/paper/2023/file/cac723e5ff29f65e3fcbb0739ae91bee-Paper-Datasets_and_Benchmarks.pdf},
 volume = {36},
 year = {2023}
}

@article{EVEREST-viruses-benchmark,
  title = {Evaluating variant effect prediction across viruses},
  url = {http://dx.doi.org/10.1101/2025.08.04.668549},
  DOI = {10.1101/2025.08.04.668549},
  publisher = {openRxiv},
  author = {Gurev,  Sarah and Youssef,  Noor and Jain,  Navami and Mehrotra,  Aarushi and Leung,  Sarrah Rose Mikhail and Jackson,  Abigail and Marks,  Debora},
  year = {2025},
  month = aug 
}

@misc{2020logitlens,
  author       = {{nostalgebraist}},
  title        = {Interpreting {GPT}: the logit lens},
  year         = {2020},
  month        = aug,
  day          = {30},
  howpublished = {\url{https://www.lesswrong.com/posts/AcKRB8wDpdaN6v6ru/interpreting-gpt-the-logit-lens}},
  note         = {LessWrong. Accessed: 2026-01-26}
}

@misc{CanonLayers,
  doi = {10.48550/ARXIV.2512.17351},
  url = {https://arxiv.org/abs/2512.17351},
  author = {Allen-Zhu,  Zeyuan},
  title = {Physics of Language Models: Part 4.1,  Architecture Design and the Magic of Canon Layers},
  publisher = {arXiv},
  year = {2025},
  copyright = {arXiv.org perpetual,  non-exclusive license}
}

@misc{gqas2JianlinSu,
  author       = {Su, Jianlin},
  title        = {{The Road to Transformer Upgrades: 20. What Makes MLA Good? (Part I)}},
  year         = {2025},
  month        = may,
  day          = {04},
  howpublished = {\url{https://kexue.fm/archives/10907}},
  note         = {Scientific Spaces. Accessed: 2026-01-25}
}

@article{colabfold,
  title = {ColabFold: making protein folding accessible to all},
  volume = {19},
  ISSN = {1548-7105},
  url = {http://dx.doi.org/10.1038/s41592-022-01488-1},
  DOI = {10.1038/s41592-022-01488-1},
  number = {6},
  journal = {Nature Methods},
  publisher = {Springer Science and Business Media LLC},
  author = {Mirdita,  Milot and Sch\"{u}tze,  Konstantin and Moriwaki,  Yoshitaka and Heo,  Lim and Ovchinnikov,  Sergey and Steinegger,  Martin},
  year = {2022},
  month = may,
  pages = {679–682}
}

@article{InterPLM,
  title = {InterPLM: discovering interpretable features in protein language models via sparse autoencoders},
  volume = {22},
  ISSN = {1548-7105},
  url = {http://dx.doi.org/10.1038/s41592-025-02836-7},
  DOI = {10.1038/s41592-025-02836-7},
  number = {10},
  journal = {Nature Methods},
  publisher = {Springer Science and Business Media LLC},
  author = {Simon,  Elana and Zou,  James},
  year = {2025},
  month = sep,
  pages = {2107–2117}
}

@misc{ProtSAE,
  doi = {10.48550/ARXIV.2509.05309},
  url = {https://arxiv.org/abs/2509.05309},
  author = {Liu,  Xiangyu and Lei,  Haodi and Liu,  Yi and Liu,  Yang and Hu,  Wei},
  title = {ProtSAE: Disentangling and Interpreting Protein Language Models via Semantically-Guided Sparse Autoencoders},
  publisher = {arXiv},
  year = {2025},
  copyright = {Creative Commons Attribution 4.0 International}
}

@article{plm-sae,
  title = {Sparse autoencoders uncover biologically interpretable features in protein language model representations},
  volume = {122},
  ISSN = {1091-6490},
  url = {http://dx.doi.org/10.1073/pnas.2506316122},
  DOI = {10.1073/pnas.2506316122},
  number = {34},
  journal = {Proceedings of the National Academy of Sciences},
  publisher = {Proceedings of the National Academy of Sciences},
  author = {Gujral,  Onkar and Bafna,  Mihir and Alm,  Eric and Berger,  Bonnie},
  year = {2025},
  month = aug 
}

@InProceedings{plm-interp-sae,
  title = 	 {From Mechanistic Interpretability to Mechanistic Biology: Training, Evaluating, and Interpreting Sparse Autoencoders on Protein Language Models},
  author =       {Adams, Etowah and Bai, Liam and Lee, Minji and Yu, Yiyang and Alquraishi, Mohammed},
  booktitle = 	 {Proceedings of the 42nd International Conference on Machine Learning},
  pages = 	 {460--476},
  year = 	 {2025},
  editor = 	 {Singh, Aarti and Fazel, Maryam and Hsu, Daniel and Lacoste-Julien, Simon and Berkenkamp, Felix and Maharaj, Tegan and Wagstaff, Kiri and Zhu, Jerry},
  volume = 	 {267},
  series = 	 {Proceedings of Machine Learning Research},
  month = 	 {13--19 Jul},
  publisher =    {PMLR},
  url = 	 {https://proceedings.mlr.press/v267/adams25a.html}
}

@article{foldseek,
  title = {Fast and accurate protein structure search with Foldseek},
  volume = {42},
  ISSN = {1546-1696},
  url = {http://dx.doi.org/10.1038/s41587-023-01773-0},
  DOI = {10.1038/s41587-023-01773-0},
  number = {2},
  journal = {Nature Biotechnology},
  publisher = {Springer Science and Business Media LLC},
  author = {van Kempen,  Michel and Kim,  Stephanie S. and Tumescheit,  Charlotte and Mirdita,  Milot and Lee,  Jeongjae and Gilchrist,  Cameron L. M. and S\"{o}ding,  Johannes and Steinegger,  Martin},
  year = {2023},
  month = may,
  pages = {243–246}
}

@inproceedings{SaProt,
 author = {Su, Jin and Han, Chenchen and Zhou, Yuyang and Shan, Junjie and Zhou, Xibin and Yuan, Fajie},
 booktitle = {International Conference on Learning Representations},
 editor = {B. Kim and Y. Yue and S. Chaudhuri and K. Fragkiadaki and M. Khan and Y. Sun},
 pages = {6987--7009},
 title = {SaProt: Protein Language Modeling with Structure-aware Vocabulary},
 url = {https://proceedings.iclr.cc/paper_files/paper/2024/file/1c42513b8895ab11fbbb5b7e8e6b6b02-Paper-Conference.pdf},
 volume = {2024},
 year = {2024}
}

@misc{progen2,
  doi = {10.48550/ARXIV.2206.13517},
  url = {https://arxiv.org/abs/2206.13517},
  author = {Nijkamp,  Erik and Ruffolo,  Jeffrey and Weinstein,  Eli N. and Naik,  Nikhil and Madani,  Ali},
  title = {ProGen2: Exploring the Boundaries of Protein Language Models},
  publisher = {arXiv},
  year = {2022},
  copyright = {Creative Commons Attribution 4.0 International}
}

@inproceedings{ProSST,
 author = {Li, Mingchen and Tan, Yang and Ma, Xinzhu and Zhong, Bozitao and Yu, Huiqun and Zhou, Ziyi and Ouyang, Wanli and Zhou, Bingxin and Tan, Pan and Hong, Liang},
 booktitle = {Advances in Neural Information Processing Systems},
 doi = {10.52202/079017-1126},
 editor = {A. Globerson and L. Mackey and D. Belgrave and A. Fan and U. Paquet and J. Tomczak and C. Zhang},
 pages = {35700--35726},
 publisher = {Curran Associates, Inc.},
 title = {ProSST: Protein Language Modeling with Quantized Structure and Disentangled Attention},
 url = {https://proceedings.neurips.cc/paper_files/paper/2024/file/3ed57b293db0aab7cc30c44f45262348-Paper-Conference.pdf},
 volume = {37},
 year = {2024}
}

@article{VenusREM,
    author = {Tan, Yang and Wang, Ruilin and Wu, Banghao and Hong, Liang and Zhou, Bingxin},
    title = {From high-throughput evaluation to wet-lab studies: advancing mutation effect prediction with a retrieval-enhanced model},
    journal = {Bioinformatics},
    volume = {41},
    number = {Supplement\_1},
    pages = {i401-i409},
    year = {2025},
    month = {07},
    issn = {1367-4811},
    doi = {10.1093/bioinformatics/btaf189},
    url = {https://doi.org/10.1093/bioinformatics/btaf189},
    eprint = {https://academic.oup.com/bioinformatics/article-pdf/41/Supplement_1/i401/63745466/btaf189.pdf},
}

@article{LOGAN,
  title = {Logan: Planetary-Scale Genome Assembly Surveys Life's Diversity},
  url = {http://dx.doi.org/10.1101/2024.07.30.605881},
  DOI = {10.1101/2024.07.30.605881},
  publisher = {openRxiv},
  author = {Chikhi,  Rayan and Lemane,  Téo and Loll-Krippleber,  Raphaël and Montoliu-Nerin,  Mercè and Raffestin,  Brice and Camargo,  Antonio Pedro and Miller,  Carson J. and Fiamenghi,  Mateus Bernabe and Agustinho,  Daniel Paiva and Majidian,  Sina and Autric,  Greg and Hugues,  Maxime and Lee,  Junkyoung and Faure,  Roland and Curry,  Kristen D. and Moura de Sousa,  Jorge A. and Rocha,  Eduardo P. C. and Koslicki,  David and Medvedev,  Paul and Gupta,  Purav and Shen,  Jessica and Morales-Tapia,  Alejandro and Sihuta,  Kate and Roy,  Peter J. and Brown,  Grant W. and Edgar,  Robert C. and Korobeynikov,  Anton and Steinegger,  Martin and Lareau,  Caleb A. and Peterlongo,  Pierre and Babaian,  Artem},
  year = {2024},
  month = jul
}

@article{OMGdataset,
  title = {The OMG dataset: An Open MetaGenomic corpus for mixed-modality genomic language modeling},
  url = {http://dx.doi.org/10.1101/2024.08.14.607850},
  DOI = {10.1101/2024.08.14.607850},
  publisher = {openRxiv},
  author = {Cornman,  Andre and West-Roberts,  Jacob and Camargo,  Antonio Pedro and Roux,  Simon and Beracochea,  Martin and Mirdita,  Milot and Ovchinnikov,  Sergey and Hwang,  Yunha},
  year = {2024},
  month = aug 
}

@misc{muon,
  author       = {Keller Jordan and Yuchen Jin and Vlado Boza and Jiacheng You and
                  Franz Cesista and Laker Newhouse and Jeremy Bernstein},
  title        = {Muon: An optimizer for hidden layers in neural networks},
  year         = {2024},
  url          = {https://kellerjordan.github.io/posts/muon/}
}

@misc{NorMuon,
  doi = {10.48550/ARXIV.2510.05491},
  url = {https://arxiv.org/abs/2510.05491},
  author = {Li,  Zichong and Liu,  Liming and Liang,  Chen and Chen,  Weizhu and Zhao,  Tuo},
  title = {NorMuon: Making Muon more efficient and scalable},
  publisher = {arXiv},
  year = {2025},
  copyright = {Creative Commons Attribution 4.0 International}
}

@misc{warmupstabledecay-lrschedule,
  doi = {10.48550/ARXIV.2410.05192},
  url = {https://arxiv.org/abs/2410.05192},
  author = {Wen,  Kaiyue and Li,  Zhiyuan and Wang,  Jason and Hall,  David and Liang,  Percy and Ma,  Tengyu},
  title = {Understanding Warmup-Stable-Decay Learning Rates: A River Valley Loss Landscape Perspective},
  publisher = {arXiv},
  year = {2024},
  copyright = {Creative Commons Attribution 4.0 International}
}

@misc{testtimescalingPLMs,
  author       = {Jung, Justin},
  title        = {Test Time Scaling of Protein Language Models},
  year         = {2026},
  month        = jan,
  day          = {08},
  howpublished = {\url{https://inference-scaling-proteins.vercel.app/}},
  note         = {Deep Exploration. Accessed: 2026-01-26}
}

@misc{cot-llm,
  doi = {10.48550/ARXIV.2201.11903},
  url = {https://arxiv.org/abs/2201.11903},
  author = {Wei, Jason and Wang, Xuezhi and Schuurmans, Dale and Bosma, Maarten and Ichter, Brian and Xia, Fei and Chi, Ed and Le, Quoc and Zhou, Denny},
  title = {Chain-of-Thought Prompting Elicits Reasoning in Large Language Models},
  publisher = {arXiv},
  year = {2022}
}

@misc{selfconsistency,
  doi = {10.48550/ARXIV.2203.11171},
  url = {https://arxiv.org/abs/2203.11171},
  author = {Wang, Xuezhi and Wei, Jason and Schuurmans, Dale and Le, Quoc and Chi, Ed and Zhou, Denny},
  title = {Self-Consistency Improves Chain of Thought Reasoning in Language Models},
  publisher = {arXiv},
  year = {2022}
}

@misc{treethoughts,
  doi = {10.48550/ARXIV.2305.10601},
  url = {https://arxiv.org/abs/2305.10601},
  author = {Yao, Shunyu and Zhao, Jeffrey and Yu, Dian and Du, Nan and Shafran, Izhak and Narasimhan, Karthik and Cao, Yuan},
  title = {Tree of Thoughts: Deliberate Problem Solving with Large Language Models},
  publisher = {arXiv},
  year = {2023}
}

@misc{tp_mup,
  doi = {10.48550/ARXIV.2203.03466},
  url = {https://arxiv.org/abs/2203.03466},
  author = {Yang,  Greg and Hu,  Edward J. and Babuschkin,  Igor and Sidor,  Szymon and Liu,  Xiaodong and Farhi,  David and Ryder,  Nick and Pachocki,  Jakub and Chen,  Weizhu and Gao,  Jianfeng},
  title = {Tensor Programs V: Tuning Large Neural Networks via Zero-Shot Hyperparameter Transfer},
  publisher = {arXiv},
  year = {2022},
  copyright = {arXiv.org perpetual,  non-exclusive license}
}

@article{ProtGPT2,
  title = {ProtGPT2 is a deep unsupervised language model for protein design},
  volume = {13},
  ISSN = {2041-1723},
  url = {http://dx.doi.org/10.1038/s41467-022-32007-7},
  DOI = {10.1038/s41467-022-32007-7},
  number = {1},
  journal = {Nature Communications},
  publisher = {Springer Science and Business Media LLC},
  author = {Ferruz,  Noelia and Schmidt,  Steffen and H\"{o}cker,  Birte},
  year = {2022},
  month = jul 
}

@article{unirep2019,
  title = {Unified rational protein engineering with sequence-based deep representation learning},
  volume = {16},
  ISSN = {1548-7105},
  url = {http://dx.doi.org/10.1038/s41592-019-0598-1},
  DOI = {10.1038/s41592-019-0598-1},
  number = {12},
  journal = {Nature Methods},
  publisher = {Springer Science and Business Media LLC},
  author = {Alley,  Ethan C. and Khimulya,  Grigory and Biswas,  Surojit S. and AlQuraishi,  Mohammed and Church,  George M.},
  year = {2019},
  month = dec,
  pages = {1315--1322}
}

@article{prottrans,
  title = {ProtTrans: Towards Cracking the Language of Life’s Code Through Self-Supervised Deep Learning and High Performance Computing},
  volume = {18},
  ISSN = {1548-7105},
  url = {http://dx.doi.org/10.1038/s41592-021-01142-2},
  DOI = {10.1038/s41592-021-01142-2},
  number = {9},
  journal = {Nature Methods},
  publisher = {Springer Science and Business Media LLC},
  author = {Elnaggar,  Ahmed and Heinzinger,  Michael and Dallago,  Christian and Rehawi,  Ghaleb and Wang,  Yu and Jones,  Llion and Gibbs,  Tom and Feher,  Tamas and Angerer,  Christoph and Steinegger,  Martin and Bhowmik,  Debsindhu and Rost,  Burkhard},
  year = {2021},
  month = sep,
  pages = {1174--1180}
}

@misc{rao2019evaluating,
  doi = {10.48550/ARXIV.1906.08230},
  url = {https://arxiv.org/abs/1906.08230},
  author = {Rao,  Roshan and Bhattacharya,  Nicholas and Thomas,  Neil and Duan,  Yan and Chen,  Xi and Canny,  John and Abbeel,  Pieter and Song,  Yun S.},
  title = {Evaluating Protein Transfer Learning with {TAPE}},
  publisher = {arXiv},
  year = {2019},
  copyright = {arXiv.org perpetual,  non-exclusive license}
}

@misc{rives2019esm,
  doi = {10.48550/ARXIV.1910.03771},
  url = {https://arxiv.org/abs/1910.03771},
  author = {Rives,  Alexander and Meier,  Joshua and Sercu,  Tom and Goyal,  Siddharth and Lin,  Zeming and Liu,  Jason and Guo,  Demi and Ott,  Myle and Zitnick,  C. Lawrence and Ma,  Jerry and Fergus,  Rob},
  title = {Biological Structure and Function Emerge from Scaling Unsupervised Learning to 250 Million Protein Sequences},
  publisher = {arXiv},
  year = {2019},
  copyright = {arXiv.org perpetual,  non-exclusive license}
}

@inproceedings{salazar2020masked,
  title        = {Masked Language Model Scoring},
  author       = {Salazar, Julian and Liang, Davis and Nguyen, Toan Q and Kirchhoff, Katrin},
  year         = {2020},
  booktitle    = {Proceedings of the 58th Annual Meeting of the Association for Computational Linguistics},
  pages        = {2699--2712}
}

@article{ragesm2025,
  title        = {RAG-ESM: Improving Pretrained Protein Language Models via Sequence Retrieval},
  author       = {Sgarbossa, Damiano and Bitbol, Anne-Florence},
  journal      = {PRX Life},
  volume       = {3},
  number       = {3},
  pages        = {033013},
  year         = {2025},
  month        = aug,
  doi          = {10.1103/db1b-hy16}
}

@inproceedings{dao2023flashattention2,
  title = {FlashAttention-2: Faster Attention with Better Parallelism and Work Partitioning},
  author = {Dao, Tri},
  booktitle = {International Conference on Learning Representations (ICLR)},
  year = {2024}
}

@misc{moddednanogpt,
  author       = {Jordan, Keller},
  title        = {modded-nanogpt: Speedrunning the NanoGPT baseline},
  year         = {2024},
  howpublished = {\url{https://github.com/KellerJordan/modded-nanogpt}},
  note         = {GitHub repository. Accessed: 2026-01-26}
}

@misc{valueresidual,
  doi = {10.48550/ARXIV.2410.17897},
  url = {https://arxiv.org/abs/2410.17897},
  author = {Zhou, Zhanchao and Wu, Tianyi and Jiang, Zhiyun and Obeid, Fares and Lan, Zhenzhong},
  title = {Value Residual Learning For Alleviating Attention Concentration In Multi-Head Attention},
  publisher = {arXiv},
  year = {2024}
}

@inproceedings{relusquared,
  title = {Searching for Efficient Transformers for Language Modeling},
  author = {So, David and Ma\'{n}ke, Wojciech and Liu, Hanxiao and Dai, Zihang and Shazeer, Noam and Le, Quoc V},
  booktitle = {Advances in Neural Information Processing Systems},
  volume = {34},
  pages = {6010--6022},
  year = {2021}
}

@misc{shazeer2020glu,
  title = {GLU Variants Improve Transformer},
  author = {Shazeer, Noam},
  year = {2020},
  eprint = {2002.05202},
  archivePrefix = {arXiv},
  primaryClass = {cs.LG}
}

@misc{deepseekv2,
  doi = {10.48550/ARXIV.2405.04434},
  url = {https://arxiv.org/abs/2405.04434},
  author = {DeepSeek-AI},
  title = {DeepSeek-V2: A Strong, Economical, and Efficient Mixture-of-Experts Language Model},
  publisher = {arXiv},
  year = {2024}
}

@misc{ropetonope,
  doi = {10.48550/ARXIV.2501.18795},
  url = {https://arxiv.org/abs/2501.18795},
  author = {Yang, Bowen and Venkitesh, Bharat and Talupuru, Dwarak and Lin, Hangyu and Cairuz, David and Blunsom, Phil and Locatelli, Acyr},
  title = {Rope to Nope and Back Again: A New Hybrid Attention Strategy},
  publisher = {arXiv},
  year = {2025}
}

@misc{decouplingwhatwhere,
  doi = {10.48550/ARXIV.2509.10534},
  url = {https://arxiv.org/abs/2509.10534},
  author = {Gopalakrishnan, Anand and Csord\'{a}s, Robert and Schmidhuber, J\"{u}rgen and Mozer, Michael C.},
  title = {Decoupling the ``What'' and ``Where'' With Polar Coordinate Positional Embeddings},
  publisher = {arXiv},
  year = {2025}
}

@misc{roundandround,
  doi = {10.48550/ARXIV.2410.06205},
  url = {https://arxiv.org/abs/2410.06205},
  author = {Barbero, Federico and Vitvitskyi, Alex and Perivolaropoulos, Christos and Pascanu, Razvan and Veli\v{c}kovi\'{c}, Petar},
  title = {Round and Round We Go! What makes Rotary Positional Encodings useful?},
  publisher = {arXiv},
  year = {2024}
}

@inproceedings{hope-vlm,
  title = {HoPE: Hybrid of Position Embedding for Long Context Vision-Language Models},
  author = {Li, Haoran and Qin, Yingjie and Ou, Baoyuan and Xu, Lai and Xu, Ruiwen},
  booktitle = {Advances in Neural Information Processing Systems},
  year = {2025}
}

@article{Gribskov1987,
  title = {Profile analysis: detection of distantly related proteins.},
  volume = {84},
  ISSN = {1091-6490},
  url = {http://dx.doi.org/10.1073/pnas.84.13.4355},
  DOI = {10.1073/pnas.84.13.4355},
  number = {13},
  journal = {Proceedings of the National Academy of Sciences},
  publisher = {Proceedings of the National Academy of Sciences},
  author = {Gribskov,  M and McLachlan,  A D and Eisenberg,  D},
  year = {1987},
  month = jul,
  pages = {4355--4358}
}

@article{bfd,
  title = {Clustering huge protein sequence sets in linear time},
  author = {Steinegger, Martin and S{\"o}ding, Johannes},
  journal = {Nature Communications},
  volume = {9},
  number = {1},
  pages = {2542},
  year = {2018},
  publisher = {Nature Publishing Group}
}

@article{mgnify,
  title = {{MGnify}: the microbiome analysis resource in 2020},
  author = {Mitchell, Alex L and Almeida, Alexandre and Beracochea, Martin and Boland, Miguel and Burgin, Josephine and Cochrane, Guy and Crusoe, Michael R and Kale, Varsha and Potter, Simon C and Richardson, Lorna J and others},
  journal = {Nucleic Acids Research},
  volume = {48},
  number = {D1},
  pages = {D570--D578},
  year = {2020},
  publisher = {Oxford University Press}
}

@article{metaeuk,
  title = {{MetaEuk}—sensitive, high-throughput gene discovery, and annotation for large-scale eukaryotic metagenomics},
  author = {Levy Karin, Eli and Mirdita, Milot and S{\"o}ding, Johannes},
  journal = {Microbiome},
  volume = {8},
  number = {1},
  pages = {48},
  year = {2020},
  publisher = {BioMed Central}
}

@article{UniRef,
  title = {{UniRef} clusters: a comprehensive and scalable alternative for improving sequence similarity searches},
  author = {Suzek, Baris E. and Wang, Yuqi and Huang, Hongzhan and McGarvey, Peter B. and Wu, Cathy H. and {UniProt Consortium}},
  journal = {Bioinformatics},
  volume = {31},
  number = {6},
  pages = {926--932},
  year = {2015},
  doi = {10.1093/bioinformatics/btu739}
}

@article{CAZy,
  title = {The carbohydrate-active enzymes database ({CAZy}): an expert resource for Glycogenomics},
  author = {Drula, Elodie and Garron, Marie-Line and Dogan, Suzan and Lombard, Vincent and Henrissat, Bernard and Terrapon, Nicolas},
  journal = {Nucleic Acids Research},
  volume = {50},
  number = {D1},
  pages = {D198--D207},
  year = {2022},
  doi = {10.1093/nar/gkab1045}
}

@article{RVDB,
  title = {{RVDB-prot}, a reference viral protein database and its {HMM} profiles},
  author = {Bigot, Thomas and Temmam, Sarah and P{\'e}rot, Philippe and Eloit, Marc},
  journal = {F1000Research},
  volume = {8},
  pages = {530},
  year = {2020},
  doi = {10.12688/f1000research.18776.2}
}

@article{VirE,
  title = {VIRE: a metagenome-derived,  planetary-scale virome resource with environmental context},
  volume = {54},
  ISSN = {1362-4962},
  url = {http://dx.doi.org/10.1093/nar/gkaf1225},
  DOI = {10.1093/nar/gkaf1225},
  number = {D1},
  journal = {Nucleic Acids Research},
  publisher = {Oxford University Press (OUP)},
  author = {Nishijima,  Suguru and Fullam,  Anthony and Schmidt,  Thomas S B and Kuhn,  Michael and Bork,  Peer},
  year = {2025},
  month = nov,
  pages = {D902–D911}
}

@article{BFVD,
  title = {BFVD—a large repository of predicted viral protein structures},
  volume = {53},
  ISSN = {1362-4962},
  url = {http://dx.doi.org/10.1093/nar/gkae1119},
  DOI = {10.1093/nar/gkae1119},
  number = {D1},
  journal = {Nucleic Acids Research},
  publisher = {Oxford University Press (OUP)},
  author = {Kim,  Rachel Seongeun and Levy Karin,  Eli and Mirdita,  Milot and Chikhi,  Rayan and Steinegger,  Martin},
  year = {2024},
  month = nov,
  pages = {D340–D347}
}

@article{Araport11,
  title = {Araport11: a complete reannotation of the {Arabidopsis thaliana} reference genome},
  author = {Cheng, Chia-Yi and Krishnakumar, Vivek and Chan, Agnes P. and Thibaud-Nissen, Francoise and Schobel, Seth and Town, Christopher D.},
  journal = {The Plant Journal},
  volume = {89},
  number = {4},
  pages = {789--804},
  year = {2017},
  doi = {10.1111/tpj.13415}
}

@article{MMseqs2,
  title = {{MMseqs2} enables sensitive protein sequence searching for the analysis of massive data sets},
  author = {Steinegger, Martin and S{\"o}ding, Johannes},
  journal = {Nature Biotechnology},
  volume = {35},
  number = {11},
  pages = {1026--1028},
  year = {2017},
  doi = {10.1038/nbt.3988}
}

@misc{RMSNorm,
  title = {Root Mean Square Layer Normalization},
  author = {Zhang, Biao and Sennrich, Rico},
  year = {2019},
  eprint = {1910.07467},
  archivePrefix = {arXiv},
  primaryClass = {cs.LG}
}

@misc{RoPE,
  title = {{RoFormer}: Enhanced Transformer with Rotary Position Embedding},
  author = {Su, Jianlin and Lu, Yu and Pan, Shengfeng and Murtadha, Ahmed and Wen, Bo and Liu, Yunfeng},
  year = {2021},
  eprint = {2104.09864},
  archivePrefix = {arXiv},
  primaryClass = {cs.CL}
}

@misc{yang2025spectral,
  doi = {10.48550/ARXIV.2310.17813},
  url = {https://arxiv.org/abs/2310.17813},
  author = {Yang,  Greg and Simon,  James B. and Bernstein,  Jeremy},
  title = {A Spectral Condition for Feature Learning},
  publisher = {arXiv},
  year = {2023},
  copyright = {Creative Commons Attribution 4.0 International}
}

@misc{polarexpress,
  doi = {10.48550/ARXIV.2505.16932},
  url = {https://arxiv.org/abs/2505.16932},
  author = {Amsel,  Noah and Persson,  David and Musco,  Christopher and Gower,  Robert M.},
  title = {The Polar Express: Optimal Matrix Sign Methods and Their Application to the Muon Algorithm},
  publisher = {arXiv},
  year = {2025},
  copyright = {arXiv.org perpetual,  non-exclusive license}
}
\bibliographystyle{icml2026}

\newpage
\appendix
\onecolumn

\section{Spectral Scaling and Learning Rate Transfer}
\label{app:spectral_ablation}

Muon replaces each gradient $G$ with its polar factor $U = G(G^\top G)^{-1/2}$, the nearest orthogonal matrix to $G$ in Frobenius norm. Since $U$ is orthogonal, all its singular values are exactly 1 and $\|U\|_* = 1$. We compute $U$ via Polar Express~\cite{polarexpress}, which solves a minimax-optimal polynomial iteration using only matrix-matrix multiplications, converging in 5 iterations in bfloat16 precision. The optimizer then rescales by $\sqrt{n_\text{out}/n_\text{in}}$, yielding $\|\Delta W\|_* = \sqrt{n_\text{out}/n_\text{in}}$. This satisfies the spectral scaling condition of~\citet{yang2025spectral}, which shows that feature learning occurs at all widths when updates obey this norm scaling, and that the learning rate can be held constant (width-independent) under this condition.

The spectral condition addresses width transfer but not depth. In standard Pre-LN transformers, the residual stream variance can grow with depth $L$, requiring the learning rate to scale inversely with depth. Sandwich normalization prevents this by applying a scaled post-norm ($1/\sqrt{L}$) to each residual branch output, bounding the signal variance independently of $L$. Concretely, each block computes $x_{\ell+1} = x_\ell + \gamma \cdot \text{RMSNorm}(F(\text{RMSNorm}(x_\ell)))$ where $\gamma = 1/\sqrt{L}$. Because the residual contribution is normalized at every layer, the optimal learning rate does not need to decrease with depth~\cite{tp_mup}. Together with Muon's width-independent spectral scaling, this allows the same learning rate to transfer across both width and depth, which is why the proxy model (12 layers, hidden 512) and the full model (24 layers, hidden 1024) share the same Muon learning rate of 0.015.

\begin{figure}[h]
\centering
\includegraphics[width=\columnwidth]{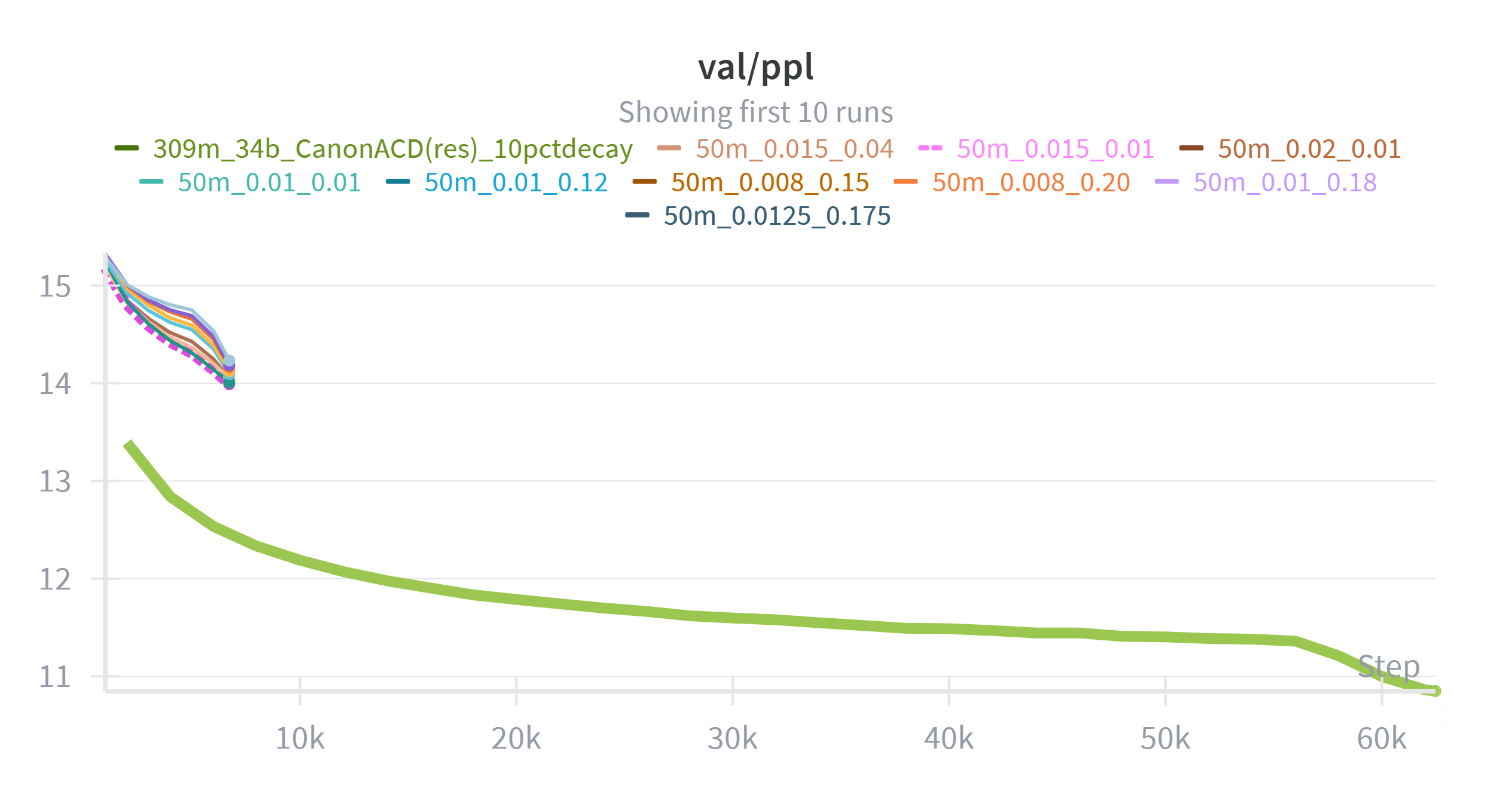}
\caption{\textbf{Learning rate transfer from 50M to 309M.} Validation perplexity curves for the 50M proxy models and the full 309M model. Legend entries are formatted as \texttt{modelsize\_learningrate\_weightdecay}.}
\label{fig:lr_transfer}
\end{figure}

We verified this empirically by training a 50M-parameter proxy model with hidden dimension 512 (half the 1024 used in Proust) on 882M tokens from the Proust training set ($\sim$0.88 Chinchilla-optimal), taking approximately 10 minutes on a single B200 GPU. The Muon learning rate of 0.015 was transferred directly to the full 309M model without retuning. Validation loss curves for the proxy and full model are shown in \Cref{fig:lr_transfer}. The transferred learning rate produces stable training and comparable convergence behavior, consistent with the theoretical findings.

\vspace{5em}

\section{Retrieval Latency}
\label{app:retrieval_latency}

ColabFold MSA retrieval latency depends on sequence length, server load, and reference databases. We measured latency across the 217 ProteinGym assays:

\begin{table}[h]
\centering
\small
\begin{tabular}{@{}lcc@{}}
\toprule
Sequence Length & Mean Latency (s) & 95th Percentile (s) \\
\midrule
$<$200 aa & 12.3 & 28 \\
200--500 aa & 24.7 & 52 \\
500--1000 aa & 38.2 & 71 \\
$>$1000 aa & 58.4 & 95 \\
\bottomrule
\end{tabular}
\caption{ColabFold MSA retrieval latency by sequence length.}
\end{table}

For batch evaluation of 217 assays, total retrieval time is approximately 4 hours (single-threaded) or 45 minutes with 8 concurrent requests. This does not include model inference time.


\end{document}